\documentclass[10pt]{article} % For LaTeX2e
\usepackage{times}
\usepackage{hyperref}
\usepackage{url}
\usepackage{algorithm}% http://ctan.org/pkg/algorithms
\usepackage{algorithmic}%
\usepackage{mathtools}
\usepackage{amsmath,amssymb,amsthm}

\title{Algebraic-Combinatorial Methods for Low-Rank Matrix Completion with Application to Athletic Performance Prediction}
\author{Duncan A.J.~Blythe\footnote{D.A.J.B and F.K. contributed equally to this work}, Louis Theran, Franz Kiraly}
\date{\today}

\theoremstyle{plain}
 \newtheorem{Thm}{Theorem}[section]
 \newtheorem{Prop}[Thm]{Proposition}

\theoremstyle{definition}
 
 \newtheorem{Def}[Thm]{Definition}



\let\phi=\varphi
\let\psi=\varpsi
\let\rho=\varrho

\begin{document}

% The \author macro works with any number of authors. There are two commands
% used to separate the names and addresses of multiple authors: \And and \AND.
%
% Using \And between authors leaves it to \LaTeX{} to determine where to break
% the lines. Using \AND forces a linebreak at that point. So, if \LaTeX{}
% puts 3 of 4 authors names on the first line, and the last on the second
% line, try using \AND instead of \And before the third author name.

\newcommand{\fix}{\marginpar{FIX}}
\newcommand{\new}{\marginpar{NEW}}

%\nipsfinalcopy % Uncomment for camera-ready version

\maketitle

\begin{abstract}
This paper presents novel algorithms which exploit the intrinsic algebraic and combinatorial structure
of the matrix completion task for estimating missing entries in the general low rank setting.
For positive data, we achieve results outperforming the state of the art nuclear norm, both in accuracy and computational
efficiency, in simulations and in the task of predicting athletic performance from partially observed
data.
\end{abstract}

\section{Introduction}

Matrix completion (MC) is the task of filling in the missing entries of a matrix. The most popular statistical model to this end is the low-rank matrix model.
Recently, an algebraic-combinatorial approach to the low rank-matrix completion problem was developed, where the authors derived the first:

\begin{enumerate}
\item algorithms which determine which entries of a matrix may be estimated exactly in the limit of low-noise, in arbitrary rank
and with \emph{any} method whatsoever \cite{KirTheTomUno12MatrixCompletion}.
\item algorithms for the rank 1 case which utilize \emph{all} information available with regard to missing entries and are thus optimal
for this case \cite{KirThe13RankOneEst}, outperforming, for example, the nuclear norm approach \cite{candes2009exact}. These algorithms work \emph{locally} on each missing
entry, leading to a fraction of the computational cost of completing all missing entries.
\item error guarantees and computable error bounds for the rank 1 case on the single entries \cite{KirThe13RankOneEst}.
\end{enumerate}

In this paper, we extend these benefits to the general low-rank setting, when the true matrix takes
positive values, as in, e.g. the \emph{Netflix} challenge. We propose algebraic-combinatorial algorithms which are

\begin{enumerate}
\item more accurate in simulations in certain regimes (high noise and low-noise, low-medium observation probability) and on real-world data.
\item considerably faster than the nuclear norm approach in completing the entire matrix.
\item \emph{even} faster when only certain entries of a matrix are required to be completed, since the algebraic-combinatorial
algorithms operate locally on the observed matrix.
\end{enumerate}

\subsection*{Related work}
Low-rank matrix completion has received a great deal of attention from the machine learning community. Three
main strands of research have developed: (1) convex relaxations of the rank constraints (e.g.,
\cite{candes2009exact, NegWai11,SalSre10,FoySre11,SreShr05}) (2) spectral
methods (e.g., \cite{KesMonOh10,Mek10,chatterjee2012matrix}), and (3) the novel algebraic approach discussed above~\cite{KirTheTomUno12MatrixCompletion,KirThe13RankOneEst}.

The approaches (1) and (2) focus on (i) estimating every missing entry; (ii) denoising every observed entry; and
(iii) minimizing the MSE over the whole matrix. The algebraic approach (3) allows for the construction of single-entry estimators which minimize the error of the entry under consideration.

\section{Theory}

\label{sec:theory}

\subsection{Positive Low Rank Model}

We assume that we observe some incomplete set of entries $E\subseteq [m]\times [n]$ of an unknown matrix $A\in \mathbb{R}^{m\times n}$ with all-positive entries.
We also assume that $A$ is a low-rank perturbation (with potentially multiplicative or additive noise) of a low rank ``true"matrix.

Positive data of this form are common in applications, e.g. the \emph{NetFlix} challenge dataset
(and in general in machine learning e.g.~motivating non-negative matrix factorization \cite{lee1999learning,fallat2000general}). This paper
presents methods which outperform the state of the art algorithms, nuclear norm \cite{candes2009exact} and OptSpace \cite{KesMonOh10}, for positive data.

\subsection{Circuits}

In this section we describe how one obtains polynomials from the observed matrix, which include variables corresponding to missing entries and which must
vanish, in limit of low noise, in order that the matrix has rank $r$. This will then yield in the following section a strategy for completing the missing entries in $A$.

The starting point here is the following classical theorem.

\begin{Thm}
Let $A\in\mathbb{C}^{m\times n}$. Then $A$ is of rank $r$ or less, if and only if all determinants of $(r+1)\times (r+1)$-submatrices of $A$ vanish. \label{thm:det_vanish}
\end{Thm}

Thus, given any submatrix of size $(r+1) \times (r+1)$, the polynomial given by its determinant with variables at its missing positions, must vanish, otherwise $A$ cannot
have rank $\leq r$.

If, this polynomial contains only one variable, $X_s$, then the polynomial may be solved uniquely, since linear in $X_s$ (the monomials in the determinant never
repeat entries). This solution may be then taken, in the noise free case as the completed entry (we approach the noisy case below).

However, in some cases, the subdeterminant may contain multiple variables. Suppose we are interested in completing just one entry $s$, i.e. solving
for $X_s$. In order to make this feasible, we need to find additional polynomials containing the remaining variables. These additional polynomials, however,
may in turn again contain additional variables. A solution for $X_s$ is thus only possible if a joint solution set may be obtained by this process
which has zero degrees of freedom. If the number of variables only proliferates with the number of polynomials added, no solution for
$X_s$ will be possible.

This observation leads us to define circuits of a given rank, which formalizes the notion of the locations on the matrix covered by this process and in particular, circuits
which allow us to complete entries in principle. The term circuit is suggestive of the fact that there is a graph theoretic interpretation to this process,
which is discussed in~\cite{KirThe13RankOneEst}; in this context, a purely algebraic formulation is possible.

First we consider collections of potential entries in the matrix which are plausible given the rank $r$ assumption

\begin{Def}
Let $S$ be a collection of indices, let $B_s\in \mathbb{C},s\in S$ be an indexed collection of numbers. Then we say that the $(B_s)_{s\in C}$ is \emph{compatible with rank $r$}, if there exists matrix $A\in \mathbb{C}^{m\times n}$, of rank $r$ or less, with $A_s=B_s$ for all $s\in C$.
\end{Def}

\begin{Def}
Let $C\subseteq [m]\times [n]$ be a subset of indices. Then $C$ is called a \emph{circuit} of rank $r$ if:
\begin{description}
\item[(i)] For every proper subset $S\subsetneq C$, any collection of numbers $B_s \in \mathbb{C},s\in S$ is compatible with rank $r$.
\item[(ii)] For any $e\in C$, and almost all collections of numbers $B_s \in \mathbb{C},s\in C\setminus\{e\},$ there are at most finitely many $B_e$ yielding a collection $B_s,s\in C$ that is compatible with rank $r$.
\end{description}
\end{Def}

This definition formalizes the notion, outlined above, that the polynomials we obtain from the matrix should at some point yield finitely many solutions for the missing entries.
The simplest circuit of rank $r$ is the support of an $(r+1)\times (r+1)$-submatrix. (i) and (ii) hold since each sub-minor with one missing entry admits exactly one compatible completion, except in a zero set of pathological cases where there are more, e.g. when all observed entries vanish.

For an $(r+1) \times (r+1)$ sub matrix, one naturally obtains a polynomial, viz. the determinant including the missing entries as variables, which vanishes
whenever the submatrix is compatible with rank $r$. It is possible to generalize this polynomial
to arbitrary circuits of rank $r$.
Thus, with each circuit, one may associate a unique \emph{circuit polynomial}, which coincides with the determinant polynomial for subminors \cite{KirTheTomUno12MatrixCompletion}.

\begin{Prop}
Let $C\subseteq [m]\times [n]$ be a circuit of rank $r$. Then, there is a(n) (up to multiplicative constant) unique irreducible polynomial $\theta_C$ in variables $X_s, s\in[m]\times [n]$ such that:
$B_s \in \mathbb{C},s\in C$ is compatible with rank $r$ if and only if $\theta_C(B_s,s\in C) = 0$.
\end{Prop}

One can prove the following theorem \cite{KirTheTomUno12MatrixCompletion} which generalizes Theorem~\ref{thm:det_vanish}:

\begin{Thm}\label{Thm:circuit}
Let $A\in\mathbb{C}^{m\times n}$, let $E\subseteq [m]\times [n]$ be a set of observed entries. Then,
the collection $A_e,e\in E$ is compatible with rank $r$ if and only if for all circuits $C\subseteq E$, the circuit polynomial evaluations $\theta_C(A_e,e\in E)$ vanish.

\end{Thm}

\section{Reconstruction by variance minimization}

\subsection{Reconstruction by circuits}
\label{sec:reconstruction}
Theorem~\ref{Thm:circuit} implies that the circuit polynomial of any circuit running through the observed entries and an unobserved entry should vanish in the low-noise limit -- thus yielding a solving strategy for that single entry.

\begin{Def}
Let $E\subseteq [m]\times [n]$ be a set of observed entries, let $e\in [m]\times [n]$ (observed or unobserved). A circuit $C\subseteq E\cup \{e\}$ of rank-$r$ with $e\in C$ is called \emph{solving circuit} for $e$ (w.r.t. $E$). If $\theta_C$ has degree $1$ in $X_e$, we call $C$ a \emph{unique solving circuit}.
\end{Def}

Every unique solving circuit gives rise to a rational \emph{solving equation} of the form \cite{KirTheTomUno12MatrixCompletion}:
$$A_e = \frac{f_C(A_s,s\in S)}{g_C(A_s,s\in S)},\quad\mbox{where}\;S=C\setminus\{e\}.$$
In the case of  a $(r+1)\times (r+1)$-submatrix, the solving equation takes the form of monomials of the determinant not containing $X_e$ on the numerator of the right hand side, and on the denominator,
monomials containing $A_e$, but with $A_e$ factored out.

In the absence of noise, it would suffice to find exactly one such equation and substitute the observed $A_s$.
If noise is present, we will follow a linear variance-minimization strategy as in the pseudocode detailed as Algorithm~\ref{alg:varmin}, which 
fulfills the desideratum that an estimate should be subject to the minimum variance when a class of estimators is under consideration.

Crucial to note at this point that a variance estimate is simultaneously an \emph{error} estimate on the prediction of \emph{individual matrix entries} we obtain. 

\begin{algorithm}
\caption{variance-minimizing local completion}

\label{alg:varmin}
\begin{algorithmic}[1]
\STATE{Find solving circuits $C_1,\dots, C_m$ for $A_{e}$}
\STATE{Compute candidate estimates $a_1,\dots, a_m$ via the $C_i$}
\STATE{Compute (co-)variance estimates $\sigma_1,\dots, \sigma_m$, from $s_e$}
\STATE{return a linear combination $\widehat{A}_{e}=\alpha_1 a_1+\dots +\alpha_m a_m$ with minimal variance}
\end{algorithmic}
\end{algorithm}

%Points which needs to be addressed are: how to find circuits $C_i$ (also, in which rank), how the estimates $a_i$ are obtained, how to obtain variance estimates from an observation model and how to determine the variance minimizing estimate.

We will present two prototypical algorithms employing this strategy: one for rank $1$ and one for general local rank $r$ matrix completion.

\subsection{Variance minimizing reconstruction: rank $1$}

For rank $1$, we use a variant of the algebraic algorithm (which we refer to as \emph{Algebraic Combinatorial Completion in Rank One} - ACCRO) presented in~\cite{KirThe13RankOneEst}. In that algorithm, computations are performed for logarithmic entries $a_e = \log A_s$, for which circuits become linear equations. In order to speed up computations, we will only consider circuits of length $4$ or less (=determinants and the entry itself, if observed) for reconstruction, also we assume that the multiplicative noise is equal on all log-entries. An additional speed-up is possible if one neglects the correlations between circuits.
We will refer to this variant as fACCRO -  \emph{fast ACCRO}. An informal description can be found in Algorithm~\ref{alg:log_regression_rank_1}. The algorithm 
follows a variance minimizing strategy as outlined in Algorithm~\ref{alg:varmin}; 
the weighting used is the logarithmic weighting of Equation~\eqref{eq:weighting}. This may in principle be used to estimate the error of the estimate.
The reason
that fACCRO is fast is that, if multiple entries are required, Step~\ref{step:store}.~may be performed \emph{en masse}.

\begin{algorithm}
\caption{fACCRO; \emph{input} incomplete matrix $A$, missing index $(i,j)$; \emph{output} completed entry $\widehat{A}_{i,j}$}
\label{alg:log_regression_rank_1}
\begin{algorithmic}[1]
\STATE{ Find all $l$ where $A(i,l)$ is observed.}
\STATE{Find all pairs $A(k,l),A(k,j)$ where both are observed.}
\STATE{Compute $w_k = |A(k,l)|$.}
\STATE{Normalize the $w_k$ so that $\sum_k w_k$ =1}
\STATE{Store $b_l = \text{exp}(\sum_{k} w_k(\text{log}(A_{k,j})-\text{log}(A_{k,l})))$ \label{step:store}}
\STATE{Estimate $\widehat{A}^k_{i,j} = A_{i,l} b_l$}
\STATE{Compute weights $w'_l = |A(i,l)|$ and normalize $\sum_k w'_l$ =1}
\STATE{Estimate $\widehat{A}_{i,j} = \text{exp}(\sum w'_l \text{log}A(i,l)$)}
\end{algorithmic}
\end{algorithm}

\subsection{Variance minimizing reconstruction: rank $r$}
\label{sec:var_min_r}

For general low rank, it is not the case that each circuit of rank $r$ determines exactly one solution when the circuit polynomial's variables
are substituted by generic entries in all but one variable $X_s$.

However, when the circuit is a unique solving circuit, and there is noise on the matrix, then $\theta_C(A_s) \approx 0$,
where $A$ is the true matrix.

In this higher rank case, the variance of the estimate given by a circuit $C$ must be approximated. To do this we perform a Taylor expansion of $\theta_C(A_s)$ around
the solution for the exact underlying matrix in Section~\ref{sec:weight_derive} of the Appendix.

The considerations of this section and the previous section lead us to define the following algorithm:
For any missing entry at position $s$, require that any determinant of a $r+1 \times r+1$ minor through that entry is 0. For each such minor $k$, estimate the variance of the estimate by Equation~\eqref{eq:weighting}, $w_s$
and the solutions given by the solving the minor determinant equations by $\hat{A}^k_s$. Then average these estimates by Algorithm~\ref{alg:varmin}, to yield an estimate of minimum variance. For large matrices, a set number of sub minors should be chosen at random for computational gains. If the observation probability is low, then one decreases $r$ until sufficient minors are present for a stable solution. See Algorithm~\ref{alg:algebraic_rank_r}.

\begin{algorithm}
\caption{vm-Closure; \emph{input} A, missing position (i,j) in almost complete $r+1 \times r+1$ subminors;\emph{output} estimate $\widehat{A}_{i,j}$ of $A_{i,j}$}
\label{alg:algebraic_rank_r}
\begin{algorithmic}[1]
\STATE{Find minors of $A$ including $A_{i,j}$ with all entries but one missing: $B^k$ for $k=1 \dots$ {\tt iterations} of size $\widehat{r}+1 \times \widehat{r}+1$ s.t. $B^k_{r+1,r+1} = A_{i,j}$}
\STATE{Set $B^{k}_0$ to be $B^{k}$ with a zero in the bottom corner, and $B^{k}_1$ a 1.}
\STATE{Set $a_1 = \text{det}(B^k_1)$ and $a_0 = \text{det}(B^k_0)$}
\STATE{Set $\delta{B_{k}} = \frac{1}{|a_1-a_0|} + \frac{|a_0|}{(a_1-a_0)^2}$}
\STATE{Define a probability measure $q(k)$ over $k$ by normalizing $\sum_{k'} \frac{1}{(\delta B_{k'})^2}\delta(k-k')$}
\STATE{Complete $\widehat{A}_{i,j} = \frac{1}{\text{\tt iterations}}\sum_{k} q(k) (1- \frac{a_0}{a_1-a_0})$}
\end{algorithmic}
\end{algorithm}

\subsection{Spectral meta-Algorithms}
\label{sec:spike}
\begin{figure}
\begin{center}
$\begin{array}{c}
\includegraphics[width=130mm,clip=true,trim= 0mm 0mm 0mm 0mm]{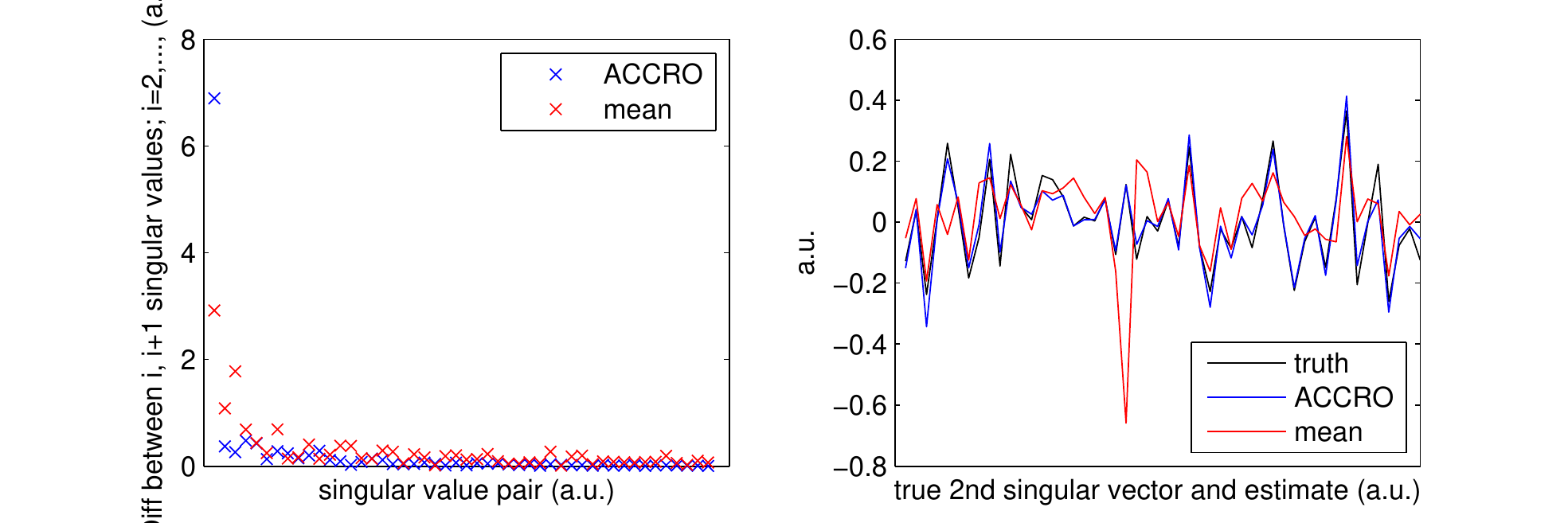}
\end{array}$
\end{center}
\caption{The figure displays in the left hand panel the difference between the 2nd and 3rd, 3rd and 4th, ... singular values on the matrix completed, by the mean (red)
and by the ACCRO algorithm (blue), from left to right, when the true matrix has rank $r=2$ and the noise level, $\epsilon = 0$, $p = 0.5$, $N=M=50$. The ACCRO exhibits a prominent separation
between the second and remaining eigenvalues, whereas, the completion by the mean, poor separation. This is again reflected in the singular vectors (right hand panel),
where the ACCRO's output (blue) well approximates the true 2nd singular vector(black) and whereas the mean completion approach approximates only poorly (red).
\label{fig:simple_example_separation}}
\end{figure}

In the example depicted in Figure~\ref{fig:simple_example_separation}, we generate a sample from model of Equation~\ref{eq:mult_noise}, in the noise free case, where the true matrix is of rank $2$.
We then apply the ACCRO algorithm to this matrix. Although this algorithm is motivated by the assumption of a rank 1 truth in its solution
strategies, we observe that the singular values (left) of the estimated matrices nevertheless reveal the rank 2 structure of the true matrix.
In the right hand panel, we see, moreover, that the second singular vector well approximates the singular true singular vector.
We compare
the same graphics (red), but for completion with the mean as per \cite{chatterjee2012matrix} ( \cite{chatterjee2012matrix} proposes a simple algorithm for matrix completion: truncation of the SVD with 0s for missing entries). We see that the singular values in this case do not reveal a clear rank 2 structure,
and the 2nd singular vector provides a poorer approximation to the true 2nd singular vector.
Our meta-Algorithms~\ref{alg:log_regression_bootstrap_rank_r} and~\ref{alg:smcb_optspace} use the fact that we may obtain the rank 2 upwards singular vectors via the algebraic algorithms in rank 1.
Both obtain the singular vectors using the output of one of the previous algorithms. The \emph{Spectral matrix completion bootstrap} (SMCB), Algorithm~\ref{alg:log_regression_bootstrap_rank_r} completes the full matrix by solving linearly for each row of $A$.
Optspace \cite{KesMonOh10} is an approach which first fills 0s into the missing entries, performs certain trimming operations, and then truncates the SVD of this coarse completion; this truncation is then fed into
an optimization routine which further refines the estimate.
Our \emph{meta-OptSpace} (mOS), Algorithm~\ref{alg:smcb_optspace}, initializes the OptSpace
optimization, instead, with the output of any of the previous algorithms. Thus we obtain algorithm instances such as mOS(SMCB(ACCRO)).%The algorithms of the previous section relied on obtaining subminors of large matrices.

These algorithms are justified by the considerable gains in accuracy over competing spectral methods such as OptSpace observed in our
simulations (Sections~\ref{sec:mc_accuracy}) and application (Section~\ref{sec:application}).
 Further
work will aim at understanding their asymptotics.

\begin{algorithm}
\caption{SMCB;  \emph{input} incomplete matrix $A$, initial estimate $A_{init}$; \emph{output} completed estimated $\widehat{A}$}
\label{alg:log_regression_bootstrap_rank_r}
\begin{algorithmic}[1]
\STATE{ Let $USV^{\top}$ be the SVD of $A_{init}$}
\FOR{i = rows of $A$}
\STATE{Let $Y = [V_1,\dots,V_r, A_{i}^\top]^\top$}
\STATE{Rearrange the rows of $Y$ to give $Y = \begin{bmatrix}A_{11}&A_{12} \\A_{21} & A_{22}\end{bmatrix}$ where $\begin{bmatrix}A_{11}&A_{12} \end{bmatrix}$ contain the singular vectors,
$A_{22}$ corresponds to the unobserved entries and $A_{21}$ the observed entries of $A_i$ ($i^\text{th}$ row of $A$)}
\STATE Set $A_{22} = A_{21}(A_{11})^+A_{12}$ where $Z^+$ is the Moore-Penrose pseudo inverse of a matrix $Z$.
\STATE{Complete the corresponding entries of $A_i$ using $A_{22}$}
\ENDFOR
\end{algorithmic}
\end{algorithm}

\begin{algorithm}
\caption{meta-OptSpace;  \emph{input} incomplete matrix $A$, initial estimate $A_{init}$; \emph{output} completed estimated $\widehat{A}$}
\label{alg:smcb_optspace}
\begin{algorithmic}[1]
\STATE{ Initialize $\widehat{A}_1$ as $A_{init}$, using any of the algebraic algorithms.}
\STATE{ Perform an SVD truncation of $\widehat{A}_1$ to rank $r$.}
\STATE Perform the optimization loop of OptSpace starting at $\widehat{A}_1$, to output $\widehat{A}$
\end{algorithmic}
\end{algorithm}

\section{Experiments on Simulated Data}

\subsection{Simulated Data}
For simulated data subject to multiplicative noise, we sample:

\begin{equation}
A = (U V^\top) \circ \mathcal{E} \label{eq:mult_noise}
\end{equation}

For simulated data subject to additive noise, we sample

\begin{equation}
A = U V^\top + \mathcal{E} \label{eq:additive_noise}
\end{equation}

Each entry of $U$, $V$ is sampled independently from $|z|$, where $z$ is a standard Gaussian. 
For multiplicative noise we consider each entry of
$\mathcal{E}$ as sampled independently from a log-normal centered around 1 ($\text{exp}(\epsilon z)$, where $\epsilon$ is the noise level).
For additive noise we consider each entry to sampled from $\epsilon |z|$.
 Each entry of the mask $M$ is sampled independently from $\{0,1\}$ from a Bernouilli distribution with parameter $p$.

\subsection{Accuracy in the Matrix Completion Task}
\label{sec:mc_accuracy}
The aim of this simulation is to assess the accuracy of the algebraic methods, baselining against
the Nuclear Norm algorithm and OptSpace.

In the first simulation,
100 matrices with additive noise and masks are realized for $\epsilon = 0.01$ and for each probability that an entry is missing, $p = 0.1,\dots,0.9$.

In the second simulation,
100 matrices with multiplicative noise and masks are realized for $\epsilon = 0.02,0.04,\dots, 0.2$ and with the probability that an entry is missing, $p = 0.6$.
In both cases compare Nuclear Norm, OptSpace, SMCB(fACCRO), vm-Closure and mOS(SMCB(fACCRO).

\begin{figure}
\begin{center}
$\begin{array}{c c}
\includegraphics[width=55mm,clip=true,trim= 0mm 0mm 0mm 0mm]{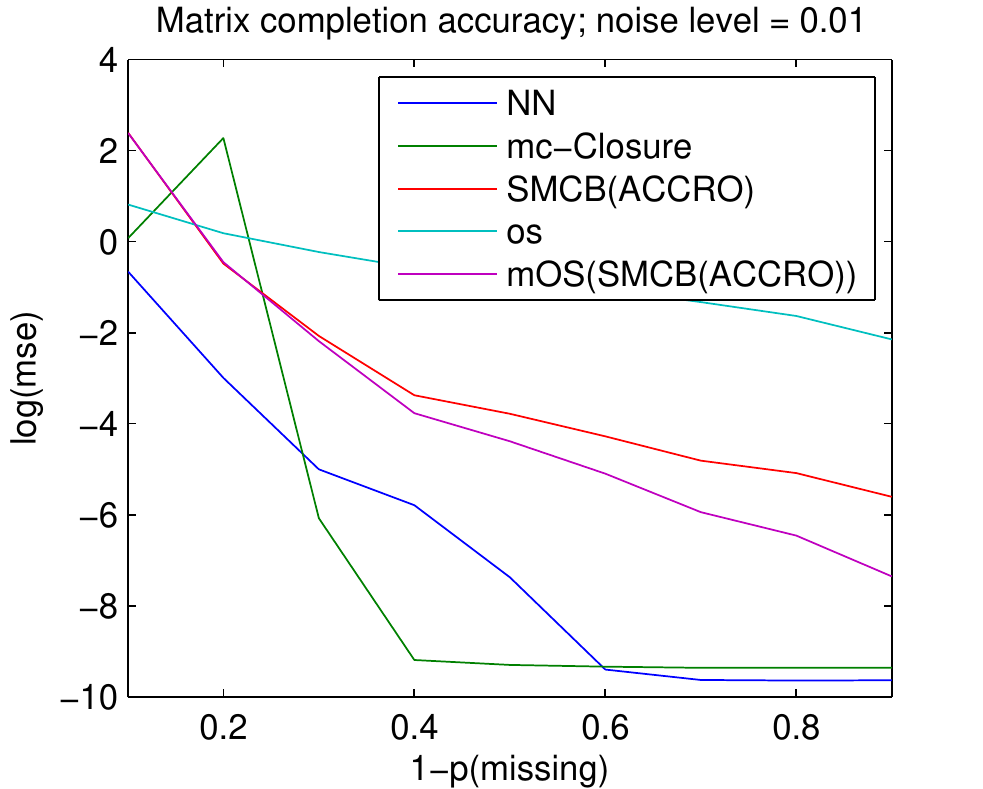} &
\includegraphics[width=55mm,clip=true,trim= 0mm 0mm 0mm 0mm]{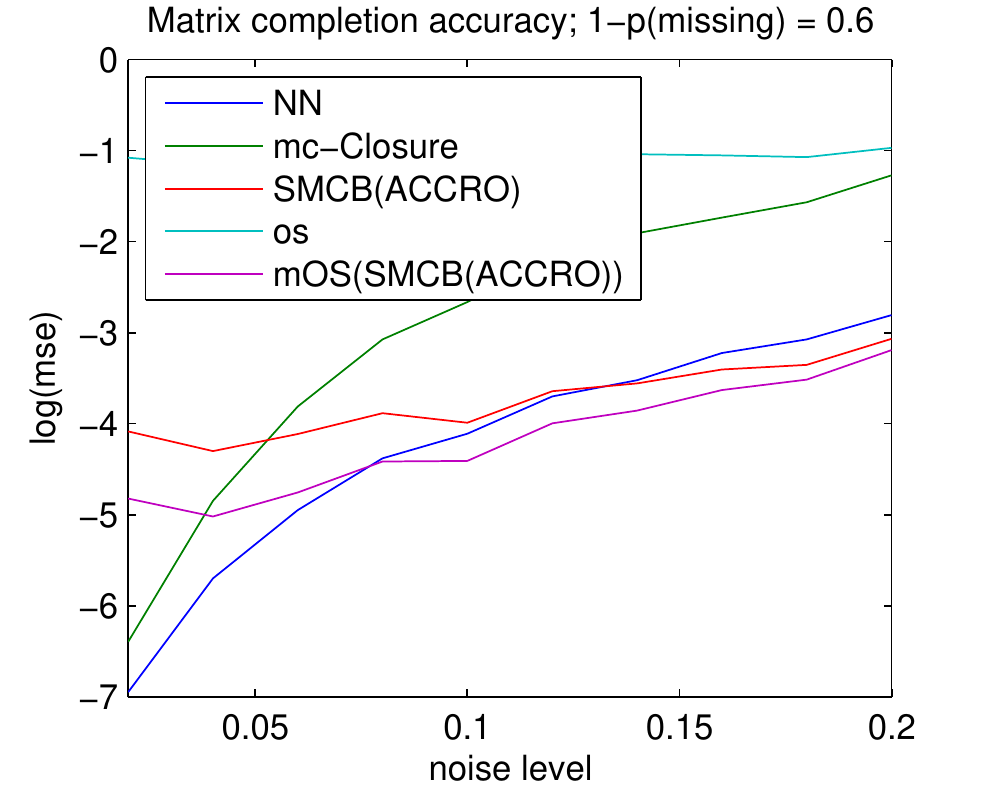} 
\end{array}$
\end{center}
\caption{The figure displays the results of the simulation described in Section~\ref{sec:mc_accuracy}. The left hand panel displays the results for additive noise.
 \label{fig:mc_accuracy}}
\end{figure}

The results are displayed in Figure~\ref{fig:mc_accuracy}. 
The left hand panel displays the results of the first simulation: the vm-Closure algorithm outperforms Nuclear at medium observation probability.
Moreover, the efficient algorithms SMCB(fACCRO) and mOS(SMCB(fACCRO) far outperform OptSpace.
The right hand panel displays the results of the second simulation. Here, the meta-algorithms, initialized by algebraic-combinatorial solutions, SMCB(fACCRO) and mOS(SMCB(fACCRO), outperform Nuclear Norm for higher noise levels;
all algebraic-combinatorial algorithms far outperform OptSpace.

\subsection{Computational Efficiency}
\label{sec:computational_efficiency}

The aim of this simulation is to compare the computational efficiency of our fastest rank $r$ algorithm, SMBC(fACCRO), with the baselines nuclear norm (with cross validation) and OptSpace.
Optspace, and our methods do not require cross validation since an estimate of the rank may be computed from the singular value spectrum.
We generate samples from the model for $p=0.5$, $\epsilon =0$ and $N = M = 20,\dots 150$ and record the computation times.
The results show that SMCB(fACCRO) is considerably more efficient than Nuclear Norm, and yields competitive accuracy; SMCB(fACCRO) outperformed in efficiency by OptSpace but provides considerably 
greater accuracy, as seen in the previous simulation.
%Due to space restrictions, we report the results in text:
%for $N=M=20$, the durations in secs. are: Nuclear Norm 4.0, SMCB(fACCRO), 0.27, OptSpace, 0.077; for $N=M=148$, Nuclear Norm 160, SMCB(fACCRO) 1.8, OptSpace 0.054.
%Thus SMCB(fACCRO) is considerably more efficient than Nuclear Norm, and yields competitive accuracy and is competitive in efficiency with OptSpace while providing considerably 
%greater accuracy.

\begin{figure}
\begin{center}
$\begin{array}{c c}
\includegraphics[width=130mm,clip=true,trim= 0mm 0mm 0mm 0mm]{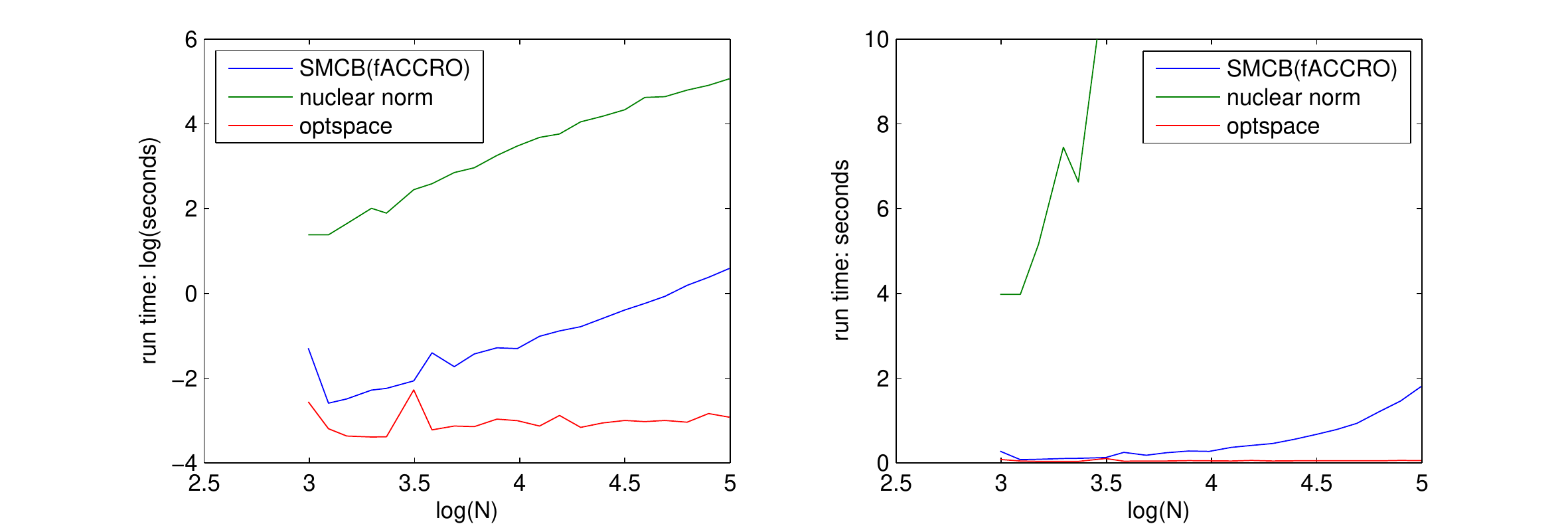}
\end{array}$
\end{center}
\caption{The figure displays the results of the simulation described in Section~\ref{sec:computational_efficiency}. The left hand panel displays computation
time in log coordinates of the tested methods (legend), Nuclear Norm, OptSpace and SMCB(fACCRO). The right hand panel displays the raw computation time.
The results show that SMCB(fACCRO) outperforms Nuclear Norm (while performing on a similar level in accuracy; see Section~\ref{sec:mc_accuracy}) whereas
although OptSpace is the fastest, it yields the poorest matrix completion performance.
 \label{fig:computational_efficiency}}
\end{figure}

\section{Application of Methods to Prediction of Athletic Performance}
\label{sec:application}
The publicly available data of a subset of runners was obtained from \url{http://www.thepowerof10.info/}, which is a database cataloguing
the performances of Great British runners, both professional and amateur. Each athlete in the database is tagged with
information on the date, location, distance as well as the performance (in hours, minutes and seconds)
over each distance.

Existing methods for the prediction of running performance have used only simple parametric models which implicitly assume that the true model has rank 1.
The best known prediction (Riegel formula) predicts linearly in log space \cite{riegel1980athletic}:
\begin{equation}
T_2 = T_1 \times (D_2/D_1)^{1.06} \label{eq:riegel}
\end{equation}
Here, $T_i$ refers to the times and $D_i$ to the corresponding distances.

We show that we can learn a higher rank model from the data which outperforms this
rank 1 method.
The existence of such a solution is tantamount to the possibility of building estimated athlete-specific information into the
predictor.

The commonly attempted distances are 100m, 200m, 400m, 800m, 1500m, 1Mile, 5000m, 10000m, Half Marathon (21.1km), Marathon (42.2km),
thus we obtain a matrix of size no.athletes $\times 10$. For each athlete $A$, we choose his/her best event relative to the population (the event for which the likelihood that another athlete $B$ is superior than $A$'s best
performance event is lowest). $A$'s best performance is entered in seconds into the corresponding column of the matrix. 
In addition, we fill in the remaining columns of the matrix with times achieved over
the remaining distances which occurred within 1 year of the best performance in $A$'s best event. The remaining entries are recorded as unobserved. In this paper we consider those athletes having attempted at least 7 events (no.athletes $\approx 400$).

Clearly the matrix takes positive values. Also important to note here is that \emph{the noise is multiplicative}. This is since, a) e.g. the expected deviation in a Marathon is on the order of minutes, whereas over 100m on the order of tenths of seconds; b)
slower runners are also more inconsistent.

\subsection{Matrix Completion Performance}

We test the MC performance, for 100 randomly deleted entries, of Nuclear Norm, OptSpace,  mOS(SMCB(ACCRO)) and mOS(SMCB(vm-Closure rank 2)); the Riegel formula (given by Equation~\eqref{eq:riegel}), serves as a baseline.
After deletion, we divide each column by the
mean of that column, so as to bring the columns on to the same scale; otherwise the mean squared error is dominated by performances over longer distances.

The results show that
a higher rank model yields a better predictor than the Riegel rank 1 predictor and that the algebraic methods significantly outperform all
baselines, including nuclear norm. Intriguing is the fact that OptSpace, when improperly initialized yields the highest MSE but when properly initialized,
the lowest. This is reminiscent of recent insights in deep learning, that an intelligent initialization of a deep network yields considerable performance gains.
\begin{center}
\begin{table}
\begin{tabular}{ c || c | c | c | c | c }
method & mOS(SMCB(ACCRO)) & nn & mOS(SMCB(vm-Cl r2)) & Riegel & OS \\
\hline
\hline
MSE $\times 10^{2}$ & {\bf 0.35} & 0.42 & {\bf 0.35} & 2.31 & 10.3  \\
 $\pm2\sigma$ & {\bf 0.019 }& 0.026 & {\bf 0.019} & 0.086 & 0.19
  \end{tabular}
\caption{Error given as MSE in dimensionless units (squared percentage of the mean time for any given distance).
The percentiles are $\pm$2 standard deviations estimated via a bootstrap with 1000 iterations.
 \label{tab:mse}}
\end{table}
\end{center}

\section{Discussion and Conclusion}

In this paper we presented algebraic combinatorial algorithms which outperform the state of the art on positive low-rank matrices in terms of accuracy and computational cost. Furthermore, the algebraic method is the only existing method which allows for the reconstruction of single entries and makes possible error estimates for them.
We conjecture that the algorithms may be generalized to non-positive or complex matrices or other incomplete data imputation tasks following a different model.

%The algorithms we present here outperform the state of the art in accuracy, in the positive matrix completion task with multiplicative noise and in efficiency.
%When the matrix is not positive, but is centered around zero, for example, the ACCRO algorithm separates estimation of the sign matrix from the
%magnitude matrix. At high noise this sign estimation introduces instabilities.
%A workaround for these difficulties is to add a positive number to every entry in the matrix, larger then the absolute value of the smallest entry expected, if known.
%This operation will increase the rank of the matrix by one.
%Additional work will revolve around extending the considerable advantages of this approach to competitiveness in all matrix completion contexts.
%
\section*{Acknowledgments}

LT is supported by the European Research Council under the European Union's Seventh Framework
Programme (FP7/2007-2013) / ERC grant agreement no 247029-SDModels.  This research was conducted
while DAJB was a guest of FK at Mathematisches Forschungsinstitut Oberwolfach, supported
by FK's Oberwolfach Leibniz Fellowship. DAJB is supported by a grant from the German Research Foundation (DFG), research training group GRK 1589/1 "Sensory Computation in Neural Systems".

%\subsubsection*{References}

\newpage

\pagestyle{empty}
\bibliographystyle{plain}
{\small
\bibliography{nips2014.bib}}

\newpage

\appendix

\section{Derivation of the weighting}

\label{sec:weight_derive}

Using the notation of Section~\ref{sec:reconstruction} one can consider the first order approximation to the estimated standard deviation of an approximate solution, as follows:
$$\delta A_e = g_C^{-1}\sum_{s\in S} f_{C,s}\cdot \delta A_s - \frac{f_C}{g_C^2} \sum_{s\in S} g_{C,s}\cdot \delta A_s,$$
where we denote $f_{C,s}=\frac{\partial f_{C}}{\partial X_s}, g_{C,s}=\frac{\partial g_{C}}{\partial X_s}$ and evaluate at $A_s$. If the noise is closer to multiplicative, considering the logarithms gives a better approximation, yielding
\begin{align*}
\delta \log A_e &= \frac{\delta A_e}{A_e} = f_C^{-1}\sum_{s\in S}f_{C,s}\cdot \delta A_s  - g_C^{-1}\sum_{s\in S} g_{C,s}\cdot \delta A_s\\
& =\frac{\delta A_e}{A_e} = f_C^{-1}\sum_{s\in S}f_{C,s}\cdot A_s \cdot \delta \log A_s  - g_C^{-1}\sum_{s\in S} g_{C,s}\cdot A_s \cdot \delta \log A_s
\end{align*}
For the determinant, both expressions take a particularly simple form. Consider an $(r+1)\times (r+1)$ matrix $A$, where all entries but the bottom right entry $A_{11}$ are observed. Write $a_k$ for the determinants of $A$ where $A_{11}$ is replaced by $k$. Note that $g_C= a_1-a_0$ and $f_C = a_0$. This yields
\begin{align*}
\delta A_{11} &=  (a_1-a_0)^{-1}\sum_{s\in S} f_{C,s}\cdot \delta A_s - \frac{a_0}{(a_1-a_0)^2} \sum_{s\in S} g_{C,s}\cdot \delta A_s,\\
\delta \log A_{11} &=  (a_0)^{-1}\sum_{s\in S} f_{C,s}\cdot \delta A_s - (a_1-a_0)^{-1} \sum_{s\in S} g_{C,s}\cdot \delta A_s\\
&=(a_0)^{-1}\sum_{s\in S} f_{C,s}\cdot A_s\cdot \delta \log A_s - (a_1-a_0)^{-1} \sum_{s\in S} g_{C,s}\cdot A_s \cdot \delta \log A_s.
\end{align*}

The sums are not easy to evaluate, even if all $\delta A_s$ are known or are of similar order of magnitude (the computation of a permanent can be obtained as a special case). However, we expect for randomly sampled data that the components of the sum lie on a single order of magnitude, therefore yielding the two approximations
\begin{align}
\delta A_{11} \approx \frac{1}{\left|a_1-a_0\right|} + \frac{\left|a_0\right|}{(a_1-a_0)^2} \quad\mbox{and}\quad \delta \log A_{11} \approx  \frac{1}{\left|a_0\right|} + \frac{1}{\left|a_1-a_0\right|} \label{eq:weighting}
\end{align}

\end{document}